\documentclass{article}
\usepackage{ijcai21}

\usepackage{times}
\usepackage{soul}
\usepackage{url}
\usepackage[hidelinks]{hyperref}
\usepackage[utf8]{inputenc}
\usepackage[small]{caption}
\usepackage{graphicx}
\usepackage{amsmath}
\usepackage{amsthm}
\usepackage{booktabs}
\usepackage{algorithm}
\usepackage{algorithmic}
\newtheorem{assumption}{Assumption}
\newtheorem{remark}{Remark}

\newcommand*\diff{\mathop{}\!\mathrm{d}}

\newcommand*\myExp{{\mathbb E}}
\newcommand*\myVar{{\mathbb V}}

\newcommand{\inputdomain}{\mathcal{X}}

\newcommand{\network}{{\cal M}}

\usepackage{centernot}
\usepackage{array}
\usepackage{multirow}
\usepackage{amsfonts}
\usepackage{xcolor}
\usepackage[utf8]{inputenc}
\usepackage{comment}
\usepackage{todonotes}
\usepackage{amsfonts, amsmath, amssymb, bbm}
\usepackage{xcolor}
\usepackage{tikz}
\usepackage{hyperref}
\usepackage{tabularx}
\usepackage{subcaption} 
\usepackage{xpatch}
\usepackage{lscape}
\usepackage[misc,geometry]{ifsym} 
\usepackage[absolute,overlay]{textpos}

\newcount\Comments  
\usepackage{color}
\definecolor{darkgreen}{rgb}{0,0.5,0}
\definecolor{purple}{rgb}{1,0,1}
\newcommand{\kibitz}[2]{\ifnum\Comments=1\textcolor{#1}{#2}\fi}

\title{Assessing the Reliability of Deep Learning Classifiers Through Robustness Evaluation and Operational Profiles}

\author{
Xingyu Zhao$^1$
\and
Wei Huang$^1$\and
Alec Banks$^2$\and
Victoria Cox$^2$\and
David Flynn$^3$\and\\
Sven Schewe$^1$
\And
Xiaowei Huang$^1$
\affiliations
$^1$University of Liverpool, Liverpool, L69 3BX, U.K.\\
$^2$Defence Science and Technology Laboratory, Salisbury, SP4 0JQ, U.K.\\
$^3$Heriot-Watt University, Edinburgh, EH14 4AS, U.K.
\emails
\{xingyu.zhao,w.huang23,sven.schewe,xiaowei.huang\}@liverpool.ac.uk \\
\{abanks,vcox\}@dstl.gov.uk,
d.flynn@hw.ac.uk
}

\begin{document}
\begin{textblock*}{20cm}(1cm,1cm)
\textcolor{red}{Preprint accepted by the AISafety'21 Workshop at IJCAI-21. To appear in a volume of CEUR Workshop Proceedings (\url{http://ceur-ws.org/})}.
\end{textblock*}
\maketitle

\begin{abstract}
The utilisation of Deep Learning (DL) is advancing into increasingly more sophisticated applications. While it shows great potential to provide transformational capabilities, DL also raises new challenges regarding its reliability in critical functions. In this paper, we present a model-agnostic reliability assessment method for DL classifiers, based on evidence from robustness evaluation and the operational profile (OP) of a given application. We partition the input space into small cells and then ``assemble'' their robustness (to the ground truth) according to the OP, where estimators on the cells' robustness and OPs are provided. 
Reliability estimates in terms of the probability of misclassification per input (pmi) can be derived together with confidence levels.
A prototype tool is demonstrated with simplified case studies. Model assumptions and extension to real-world applications are also discussed. While our model easily uncovers the inherent difficulties of assessing the DL dependability (e.g.\ lack of data with ground truth and scalability issues), we provide preliminary/compromised solutions to advance in this research direction.
\end{abstract}

\section{Introduction}
\label{sec_intro}

Industry is adopting increasingly more advanced big data analysis methodologies to enhance the operational performance, safety, and lifespan of their products and services. For many products and systems high in-service reliability and safety are key targets to ensure customer satisfaction and regulatory compliance, respectively. AI and Deep Learning (DL) have steadily grown in interest and applications. Key industrial foresight reviews have identified that the biggest obstacle to reap the benefits of DL-powered robots is the assurance and regulation of their safety and reliability \cite{lane_new_2016}.
Thus, there is an urgent need to develop methods to enable the dependable use of AI/DL in critical applications \cite{robu_train_2018} and, more importantly, to \textit{assess} and 
\textit{demonstrate} the dependability for certification and regulation.

For traditional systems, safety and reliability analysis is guided by established standards, and supported by mature development processes and verification and validation (V\&V) tools and techniques.
The situation is different for systems that utilise DL: they require new and advanced analysis reflective of the complex requirements in their safe and reliable function.
Such analysis also needs to be tailored to fully evaluate the inherent character of DL \cite{BKCF2019}, despite the progress made recently \cite{huang_survey_2020}.

DL classifiers are subject to robustness concerns, reliability models without considering robustness evidence are not convincing. Reliability, as a user-centred property, depends on the end-users' behaviours \cite{littlewood_software_2000}.
The operational profile (OP) information (quantifying how the software will be operated \cite{musa_operational_1993}) should therefore be explicitly modelled in the assessment. However, to the best of our knowledge, there is no dedicated reliability assessment model (RAM) taking into account both the OP and robustness evidence, which motivates this research.

In \cite{zhao_safety_2020}, we propose a safety case framework tailored for DL, in which we describe an initial idea of combining robustness verification and operational testing for reliability claims.
In this paper, we implement this idea as a RAM, inspired by partition-based testing \cite{hamlet_partition_1990}, operational-profile testing \cite{strigini_guidelines_1997,zhao_assessing_2020} and DL robustness evaluation \cite{carlini_towards_2017,webb_statistical_2019}. It is \textit{model-agnostic} and designed for \textit{pretrained} DL models, yielding upper bounds on the \textit{probability of miss-classifications per input} (\textit{pmi})\footnote{This reliability measure is similar to the conventional probability of failure per demand (\textit{pfd}), but retrofitted for classifiers.} with confidence levels.
Although our RAM is theoretically sound, we discover some issues in our case studies (e.g.\ scalability and lack of data) that we believe represent the \textit{inherent difficulties} of assessing/assuring DL dependability.  

The key contributions of this work are:

\textit{a)} A first RAM for DL classifiers based on the OP information and robustness evidence.

\textit{b)} Discussions on model assumptions and extension to real-world applications, highlighting the inherent difficulties of assessing DL dependability uncovered by our model.

\textit{b)} A prototype tool\footnote{Available at \url{https://github.com/havelhuang/ReAsDL}.} of our RAM with preliminary and compromised solutions to those uncovered difficulties.

\paragraph{Related Work}
\label{sec_related_work}
In recent years, there has been extensive efforts in verifying DL robustness, evaluating generalisation errors, and detecting adversarial examples (AEs). They are normally based on formal methods \cite{huang_safety_2017,katz_marabou_2019} or statistical approaches \cite{webb_statistical_2019,weng_proven_2019}. 
A comprehensive review of those techniques can be sourced from recent survey papers \cite{huang_survey_2020,zhang_machine_2020}. To the best of our knowledge, the only papers on testing DL for assessment within an operational context are \cite{li_boosting_2019,guerriero_operation_2021}. In \cite{li_boosting_2019}, novel stratified sampling methods are used to improve the operational testing efficiency. Similarly, \cite{guerriero_operation_2021} presents a sampling method from the operational dataset leveraging ``auxiliary information for misclassification'', so that it provides unbiased statistical assessment while exposing as many misclassifications as possible. However, neither of them considers robustness evidence in their assessment models. 


At the higher level of whole-systems utilising DL, although there are RAMs based on operational data, knowledge from low-level DL components is usually ignored, e.g., \cite{kalra_driving_2016}. In \cite{zhao_assessing_2020}, we improved \cite{kalra_driving_2016} by providing a Bayesian mechanism to combine such knowledge, but did not show where to obtain the knowledge. In that sense, this paper is also a follow up of \cite{zhao_assessing_2020}, forming the prior knowledge required. 



\paragraph{Organisation of the paper} We first present preliminaries on OP-based software reliability assessment and DL robustness. Then Section~\ref{sec_the_ram} describes the RAM in details with a running example. We conduct case studies in Section~\ref{sec_evaluation}, while discuss the model assumptions and extensions in Section~\ref{sec_discussion}. Finally, we conclude in Section~\ref{sec_conclusion} with future work.

\section{Preliminaries}
\label{sec_preliminary}

\subsection{OP Based Software Reliability Assessment}

The \textit{delivered reliability}, as a \textit{user-centred} and \textit{probabilistic} property, requires to model the end-users' behaviours (in the running environments) and to be formally defined by a quantitative metric \cite{littlewood_software_2000}.
Without loss of generality, we focus on \textit{pmi} as a generic metric for DL classifiers, where inputs are, e.g., facial images uploaded by users for facial recognition. We discuss later how \textit{pmi} can be redefined to cope with real-world applications like traffic sign detection.
If we denote the unknown \textit{pmi} as a variable $\lambda$, then
\begin{equation}
\label{eq_pfd_def}
\lambda:=\int_{x\in \inputdomain} I_{\{x \text{ causes a misclassification}\}}(x)Op(x)\diff{x}
\end{equation}
where $x$ is an input in the input domain\footnote{We assume continuous $\inputdomain$ in this paper. For discrete $\inputdomain$, the integral in Eq.~\eqref{eq_pfd_def} reduces to sum and OP is a probability mass function.} $\inputdomain$, and ${I}_{\tt S}$ is an indicator function---it is equal to 1 when {\tt S} is true and 0 otherwise.
The $Op(x)$ 
returns the probability that $x$ is the next random input, the OP \cite{musa_operational_1993}, a notion used in software engineering to quantify how the software will be operated. Mathematically, the OP is a probability density function (PDF) defined over $\inputdomain$.

Assuming independence between successive inputs defined in our \textit{pmi}, we may use the Bernoulli process
as the mathematical abstraction of the failure process (common for such ``on-demand'' type of systems), which implies a Binomial likelihood. Normally for traditional software, upon establishing the likelihood, RAMs on estimating $\lambda$ vary case by case---from the basic Maximum Likelihood Estimation (MLE) to Bayesian estimators tailored for certain scenarios when, e.g., seeing no
failure \cite{bishop_toward_2011}, inferring ultra-high reliability \cite{zhao_assessing_2020}, with certain forms of prior knowledge like perfectioness \cite{strigini_software_2013}, and with vague prior knowledge that expressed in imprecise probabilities \cite{walter_imprecision_2009,zhao_probabilistic_2019}.

OP based RAMs designed for traditional software fail to consider new characteristics of DL, e.g., unrobustness and high-dimensional input space. Specifically, it is quite hard to have the required prior knowledge in those Bayesian RAMs. While frequentist RAMs would require a large sample size to gain enough confidence in the estimates due to the extremely large population size (high-dimensional pixel space), especially for a high-reliable DL model where misclassifications are rare-events. As an example, the usual accuracy testing of DL classifiers is essentially an MLE estimate against the test set. It not only assumes the test set statistically represents the OP (our Assumption \ref{assumption_dataset_represents_OP} later), but also requires a large number of samples to claim high reliability with sufficient confidence.


\subsection{DL Robustness and the $R$-Separation Property}


DL is known not to be robust. Robustness requires that the decision of the DL model $\network$ is invariant against small perturbations on inputs. 
That is, all inputs in a region $ \eta \subset \inputdomain$ have the same prediction label, where usually the region $\eta$ is a small norm ball (in a $L_{p}$-norm distance\footnote{Distance mentioned in this paper is defined in $L_\infty$.}) of radius $\epsilon$ around an input $x$.
Inside $\eta$, if an input $x'$ is classified differently to $x$ by $\network$, then $x'$ is an AE.
Robustness can be defined either as a binary metric (if there exists any adversarial example in $\eta$) or as a probabilistic metric (how likely the event of seeing an adversarial example in $\eta$ is).
The former aligns with formal verification, e.g.\ \cite{huang_safety_2017}, while the latter is normally used in statistical approaches, e.g. \cite{webb_statistical_2019}. The former ``verification approach'' is the binary version of the latter ``stochastic approach''\footnote{Thus, we use the more general term robustness ``evaluation'' rather than robustness ``verification'' throughout the paper.}.

Similar to \cite{webb_statistical_2019}, we adopt the more general probabilistic definition on the robustness of the model $\network$ (in a region $\eta$ and to a target label $y$):
{\small
\begin{equation}
\label{eq_robust_def}
R_{\network}(\eta, y):=\sum_{x \in \eta} I_{\{\network(x) \text{ predicts label } y \}}(x)\times Op(x\mid x \in \eta)
\end{equation}
}\normalsize
where $Op(x \mid x\in\eta)$ is the \textit{conditional OP} of region $\eta$ (precisely the ``input model'' defined in \cite{webb_statistical_2019} and also used in \cite{weng_proven_2019}).

We highlight the follow two remarks regarding robustness:
\begin{remark}[astuteness]
\label{remark_astutenss}
Reliability assessment only concerns the robustness to the ground truth label, rather than an arbitrary label $y$ in $R_{\network}(\eta, y)$. When $y$ is such a ground truth, robustness becomes \textbf{astuteness} \cite{yang_closer_2020}, which is also the \textbf{conditional reliability} in the region $\eta$.
\end{remark} 
Astuteness is a special case of robustness\footnote{Thus, later in this paper, we may refer robustness to astuteness for brevity when it is clear from the context.}.
An extreme example showing why we introduce the concept of astuteness is:
a perfectly robust classifier that always outs ``dogs'' for any given input is unreliable. Thus, robustness evidence cannot directly support reliability claims unless the ground truth label is used in $R_{\network}(\eta, y)$.
\begin{remark}[$r$-separation]
\label{remark_r_sep}
For real-world image datasets, any data-points with different ground truth are at least distance $2r$ apart in the input space $\inputdomain$ (i.e., pixel space), and $r$ is bigger than usual norm ball radius in robustness studies.
\end{remark}

\begin{figure}[ht]
	\centering
	\includegraphics[width=\linewidth]{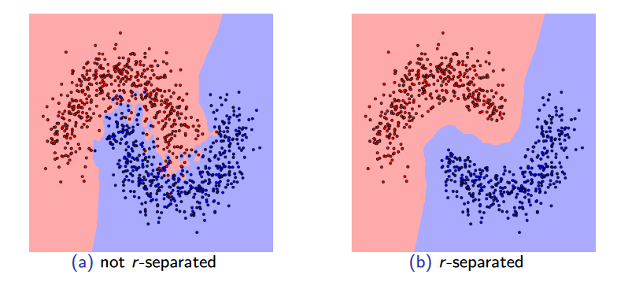}
	\caption{Illustration of the $r$-separation property.}
	\label{fig_r_sep_demo}
\end{figure}

The $r$-separation property was first observed by \cite{yang_closer_2020}: real-world image datasets studied by the authors implies that $r$ is normally $3\sim 7$ times bigger than the radius (denoted $\epsilon$) of norm balls commonly used in robustness studies. Intuitively it says that, although the classification boundary is highly non-linear, there is a minimum distance between two real-world objects of different classes (cf.\ Figure~\ref{fig_r_sep_demo} for a conceptual illustration). Moreover, such minimum distance is bigger than the usual norm ball size in robustness studies. 




\section{A RAM for Deep Learning Classifiers}
\label{sec_the_ram}

\subsection{The Running Example}
\label{sec_running_exp}
\begin{figure}[ht]
	\centering
	\includegraphics[width=\linewidth]{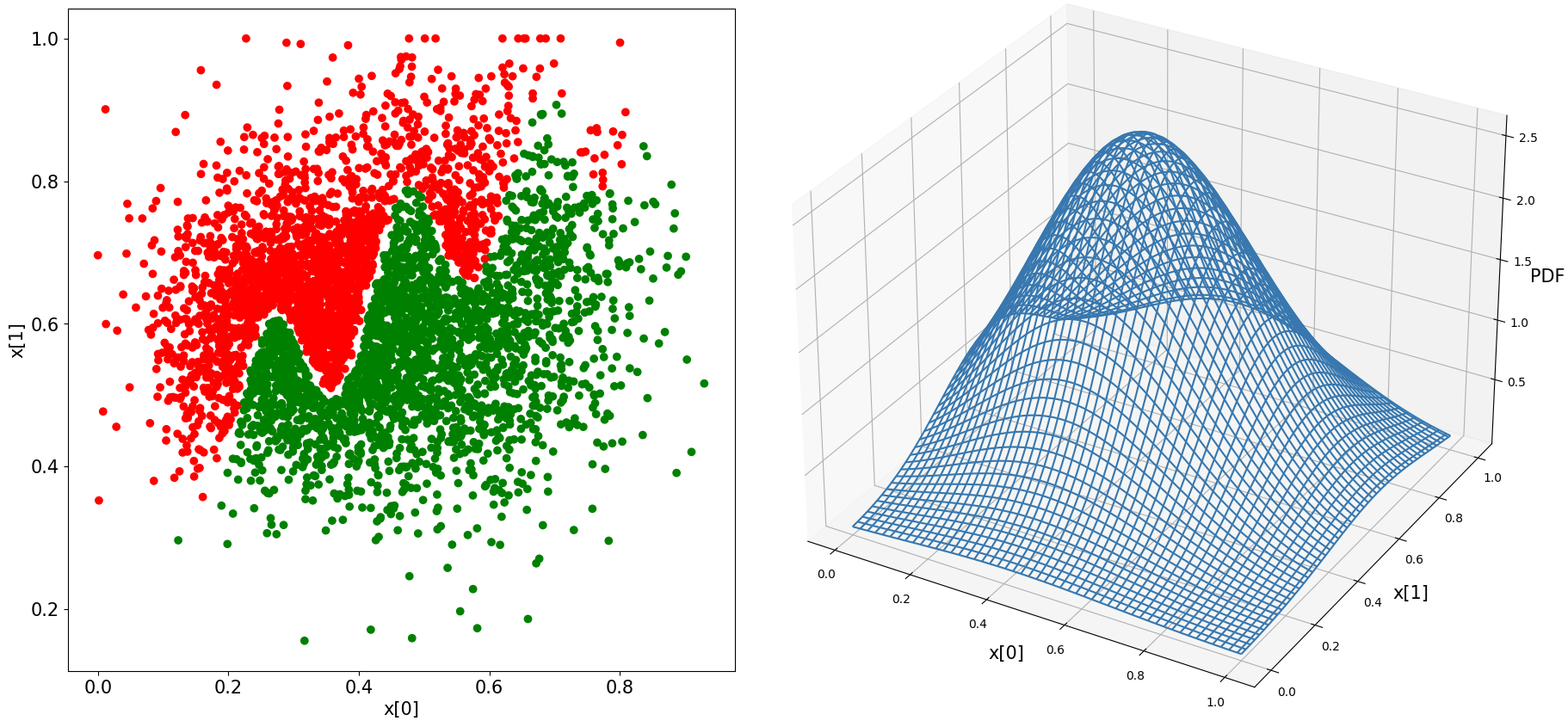}
	\caption{The 2D-point dataset (lhs), and its approximated OP (rhs).}
	\label{fig_running_example}
\end{figure}

To better demonstrate our RAM, we take the Challenge of AI Dependability Assessment raised by the Siemens Mobility\footnote{\url{https://ecosystem.siemens.com/ai-da-sc/}} as a running example. Basically, the challenge is to firstly train a DL model to classify a dataset generated on the unit square $[0,1]^2$ according to some unknown distribution. The collected data-points (training set) are shown in Figure~\ref{fig_running_example}~(lhs). Then we need to build a RAM to claim an upper bound on the probability that the next random point is miss-classified, i.e.\ \textit{pmi}. If the 2D-points represent traffic lights, then we have 2 types of misclassifications---safety-critical ones when red data-point is labelled green, and performance related otherwise. For brevity, we only focus on misclassifications here, while our RAM can cope with sub-types of misclassifications.

\subsection{The Model}
\label{sec_the_model}

\paragraph{The Framework} 
Inspired by \cite{pietrantuono_reliability_2020}, the general idea of our RAM is to partition the input domain into $m$ small cells, subject to the $r$-separation property. Then, for each cell $c_i$ (with a single ground truth $y_i$), we estimate:
\begin{align}
\label{eq_cell_rel_op}
\lambda_i&:=1-R_{\network}(c_i, y_i) \mbox{ and }
Op_i:=\sum_{x\in c_i}Op(x)
\end{align}
which are the \textit{unastuteness} and \textit{pooled OP}, respectively, estimates of the cell $c_i$---we introduce estimators for both later. Then Eq.~\eqref{eq_pfd_def} can be written as the weighted sum of the \textit{cell-wise} unastuteness (i.e.\ the conditional \textit{pmi} of each cell\footnote{We use ``cell unastuteness'' and ``cell \textit{pmi}'' interchangeably later.}) where the weights are the pooled OP of cells: 
\begin{equation}
\label{eq_pfd_cell}
\lambda=\sum_{i = 1}^{m} Op_i \lambda_i  
\end{equation}

Eq.~\eqref{eq_pfd_cell} represents an ideal case in which we know those $\lambda_i$s and $Op_i$s with certainty. In practice, we can only estimate them with imperfect estimators yielding, e.g., a point estimate with variance capturing the measure of trust. To propagate the confidence in the estimates of $\lambda_i$s and $Op_i$s, we assume:
\begin{assumption}
\label{assumption_op_lambda_indep}
All $\lambda_i$s and $Op_i$s are independent unknown variables under estimations.
\end{assumption}
\noindent Then, the estimate of $\lambda$ and its variance are:
{\small
\begin{align}
\label{eq_expected_pfd}
\myExp[\lambda] &= \sum_{i = 1}^m \myExp[\lambda_i Op_i]= \sum_{i = 1}^m \myExp[\lambda_i] \myExp[Op_i]
\\
\myVar[\lambda] &= \sum_{i=1}^m \myVar[\lambda_i Op_i]\nonumber
\\
&= \sum_{i=1}^m \myExp[\lambda_i]^2 \myVar{[Op_i]} + \myExp[Op_i]^2 \myVar[\lambda_i]
 + \myVar[\lambda_i] \myVar[Op_i]
 \label{eq_varaince_pfd}
\end{align}
}\normalsize
Note that, for the variance, the covariance terms are dropped out due to the independence assumption.

Depending on the specific estimators adopted, certain parametric families of the distribution of $\lambda$ can be assumed, from which any quantile of interest (e.g.\ 95\%) can be derived as our confidence bound in reliability. For instance, as readers will see later, we may assume $\lambda \sim \mathcal{N}(\myExp[\lambda], \myVar[\lambda])$ since all $\lambda_i$s and $Op_i$s are normal distributed variables after applying the Central Limit Theorem (CLT) in our chosen estimators.
Then, an upper bound with $1-\alpha$ confidence is
\begin{equation}
\label{eq_pfd_ci}
\mathit{Ub}_{1-\alpha} =  \myExp[\lambda] + z_{1-\alpha} \sqrt{\myVar[\lambda]}
\end{equation}
where $Pr(Z \leq z_{1-\alpha}) = 1-\alpha$, and $Z \sim \mathcal{N}(0,1)$ is a standard normal distribution.

Now the the problem is reduced to how to obtain the estimates $\myExp[\lambda_i]$s and $\myVar[\lambda_i]$s, for which we will discuss as follows referring to the running example.

\paragraph{Partition of the Input Domain $\inputdomain$} As per Remark \ref{remark_astutenss}, the astuteness evaluation of a cell requires its ground truth label. To leverage the $r$-separation property and Assumption \ref{assumption_single_gt_cell}, we partition the input space by choosing a cell radius $\epsilon$ so that $\epsilon < r$. Although we concur with Remark \ref{remark_r_sep} (first observed by \cite{yang_closer_2020}) and believe that there should exist an \textit{$r$-stable ground truth} (which means that the ground truth is stable in such a cell)
for any real-world DL classification applications, it is hard to estimate such an $r$ (denoted by $\hat{r}$) and the best we can do is to assume:
\begin{assumption}
\label{assumption_r_estiamtes_from_data}
There is a $r$-stable ground truth
(as defined in Remark \ref{remark_r_sep}) for any real-world classification problems, and it can be sufficiently estimated from the existing dataset.
\end{assumption}

That said, we get $\hat{r}=0.004013$ by iteratively calculating the minimum  distance of different labels in the running example. Then we choose a cell radius\footnote{Radius in $L_{\infty}$ which is the side length of our square cell in $L_2$.} $\epsilon=0.004$ and partition the unit square $\inputdomain$ into $250 \times 250$ cells.

\paragraph{Cell OP Approximation}
Given a dataset $(X,Y)$, we estimate the pooled OP of cell $c_i$ to get $\myExp[Op_i]$ and $\myVar[Op_i]$. We use the well-established Kernel Density Estimation (KDE) to fit a $\widehat{Op}(x)$ to approximate the OP. 
\begin{assumption}
\label{assumption_dataset_represents_OP}
The existing dataset $(X,Y)$ are randomly sampled from the OP, thus statistically represents the OP.
\end{assumption}
\noindent This assumption may not hold in practice: training data is normally collected in a \textit{balanced} way, since the DL model is expected to perform well in all categories of inputs, especially when the OP is unknown at the time of training and/or expected to change in future. Although our model can relax this assumption (cf. Section~\ref{sec_discussion}), we adopt it for brevity in demonstrating the running example. 

Then given a set of (unlabelled) data-points $(X_1,\dots, X_n)$ from the existing dataset $(X,Y)$, KDE yields
\begin{equation}
\label{eq_kde}
\widehat{Op}(x) = \frac{1}{n h} \sum_{j = 1}^{n} K(\frac{x-X_j}{h})
\end{equation}
where $K$ is the kernel function (e.g. Gaussian or exponential kernels), and $h > 0$ is a smoothing parameter called the bandwidth, cf. \cite{silverman1986density} for guidelines on tuning $h$. The approximated OP\footnote{With a Gaussian kernel and $h=0.2$ that optimised by cross-validated grid-search \cite{bergstra_random_2012}.} is shown in Figure~\ref{fig_running_example}~(rhs). 

Since our cells are small and all equal size, instead of calculating $\int_{x\in c_i}\widehat{Op}(x)dx$, we may approximate $Op_i$ as
\begin{equation}
\label{eq_cell_op}
\widehat{Op}_i = \widehat{Op}\left(x_{c_i}\right) v_c
\end{equation}
where $\widehat{Op}(x_{c_i})$ is the probability density at the cell's central point $x_{c_i}$, and $v_c$ is the constant cell volume ($1.6e{-5}$ in the running example).


Now if we introduce new variables $W_j = \frac{1}{h} K(\frac{x-X_j}{h})$, the KDE evaluated at $x$ is actually the sample mean of $W_1,\dots,W_n$.
Then by CLT, we have $\widehat{Op}(x) \sim \mathcal{N}(\mu_W,\frac{\sigma_W^2}{n})$ where the mean and variance of $\widehat{Op}(x)$ are known results:
\begin{align}
\label{eq_clt_KDE_mean}
\myExp[\widehat{Op}(x)] &= \frac{1}{nh} \sum_{j = 1}^{n} K(\frac{x-X_j}{h})
\\
\myVar[\widehat{Op}(x)] &= \frac{f(x) \int K^2(u) du}{nh} + O(\frac{1}{nh}) \approx \hat{\sigma}^2_B(x)
\label{eq_clt_KDE_var}
\end{align}
where the last step of Eq.~\eqref{eq_clt_KDE_var} says that $\myVar[\widehat{Op}(x)]$ can be approximated using a bootstrap variance $\hat{\sigma}^2_B(x)$ \cite{chen2017tutorial} (cf. the Appendix A for details).

Upon establishing Eq.s~\eqref{eq_clt_KDE_mean} and \eqref{eq_clt_KDE_var}, together with Eq.~\eqref{eq_cell_op}, we know for a given cell $c_i$ (knowing its central point $x_{c_i}$):
\begin{align}
\myExp[Op_i]=v_{c}\myExp[\widehat{Op}(x_{c_i})],\quad
\myVar[Op_i]=v_{c}^2\myVar[\widehat{Op}(x_{c_i})]
\end{align}
which are the cell OP estimates for Eq.s~\eqref{eq_expected_pfd} and \eqref{eq_varaince_pfd}.

\paragraph{Cell Astuteness Evaluation}

As a corollary of Remark \ref{remark_r_sep} and Assumption \ref{assumption_r_estiamtes_from_data}, we may confidently assume:
\begin{assumption}
\label{assumption_single_gt_cell}
If the radius of $c_i$ is smaller than $r$, all data-points in the region $c_i$ share a single ground truth label.
\end{assumption}

Now, to determine the ground truth label of a cell $c_i$, we can classify our cells into three types:

\textit{a)} Normal cells: a normal cell contains data-points sharing a same ground truth label, which is then determined as the ground truth label of the cell.

\textit{b)} Empty cells:
a cell is ``empty'' in the sense that no data-point that has been observed in it. Due to the lack of data, it is hard to determine an empty cell's ground truth. For now, we do voting based on the predicted labels (by the DL model) of random samples from the cell, assuming:
\begin{assumption}
\label{assumption_empty_cell_label}
The accuracy of the DL model is better than a classifier doing random classifications in any given cell.
\end{assumption}
\noindent Essentially the above assumption relates to the oracle problem of DL testing, that we see some recent efforts, e.g. \cite{guerriero_reliability_2020}, may relax it.

\textit{c)} Cross-boundary cells: our estimate on $r$ is imperfect, thus we may still observe data-points with different labels in one cell. Such cells are crossing the classification boundary. If our estimate on $r$ is sufficiently accurate, they should be very rare. Thus, without the need to determine the ground truth label of a cross boundary cell, we simply and conservatively set the cell unastuteness to 1.

So far, the problem is reduced to: given a normal or empty cell $c_i$ with the known ground truth label $y_i$, evaluate the miss-classification probability upon a random input $x \in c_i$, i.e.\ $\myExp[{\lambda_i}]$ and its variance $\myVar[{\lambda_i}]$. This is essentially a statistical problem that has been studied in \cite{webb_statistical_2019} using Multilevel Splitting Sampling, while we use the Simple Monte Carlo method for brevity in the running example:
$$
\hat{\lambda}_i = \frac{1}{n} \sum_{j = 1}^n I_{\{M(x_j) \neq y_i\}}
$$
The CLT tells us $\hat{\lambda}_i \sim \mathcal{N}(\mu, \frac{\sigma^2}{n})$, when $n$ is large, where $\mu$ and $\sigma^2$ are population mean and variance of $I_{\{\network(x_j) \neq y_i\}}$ that can be approximated with sample mean $\hat{\mu}_n$ and sample variance $\hat{\sigma}_{n}^2/n$. Finally, we can get
\begin{align}
\myExp[{\lambda_i}]&= \hat{\mu}_n = \frac{1}{n} \sum_{j = 1}^n I_{\{\network(x_j) \neq y_i\}} 
\\ 
\myVar[{\lambda_i}] &= \frac{\hat{\sigma}_{n}^2}{n} = \frac{1}{(n-1)n} \sum_{j = 1}^n (I_{\{\network(x_j) \neq y_i\}} - \hat{\mu}_{n})^2
\end{align}

Notably, to solve the above statistical problem with sampling methods, we need to assume how the inputs in the cell are distributed, i.e., a distribution for the conditional OP $Op(x \mid x\in c_i)$. Without loss of generality, we assume:
\begin{assumption}
\label{assumption_conditonal_OP_uniform}
The inputs in a small region like cells are uniformly distributed.
\end{assumption}
\noindent which is not uncommon (e.g., in \cite{webb_statistical_2019,weng_proven_2019}) and can be easily replaced by other distributions if there is supporting evidence for such action.

\section{Case Studies}

\begin{figure}[ht]
	\centering
	\includegraphics[width=\linewidth]{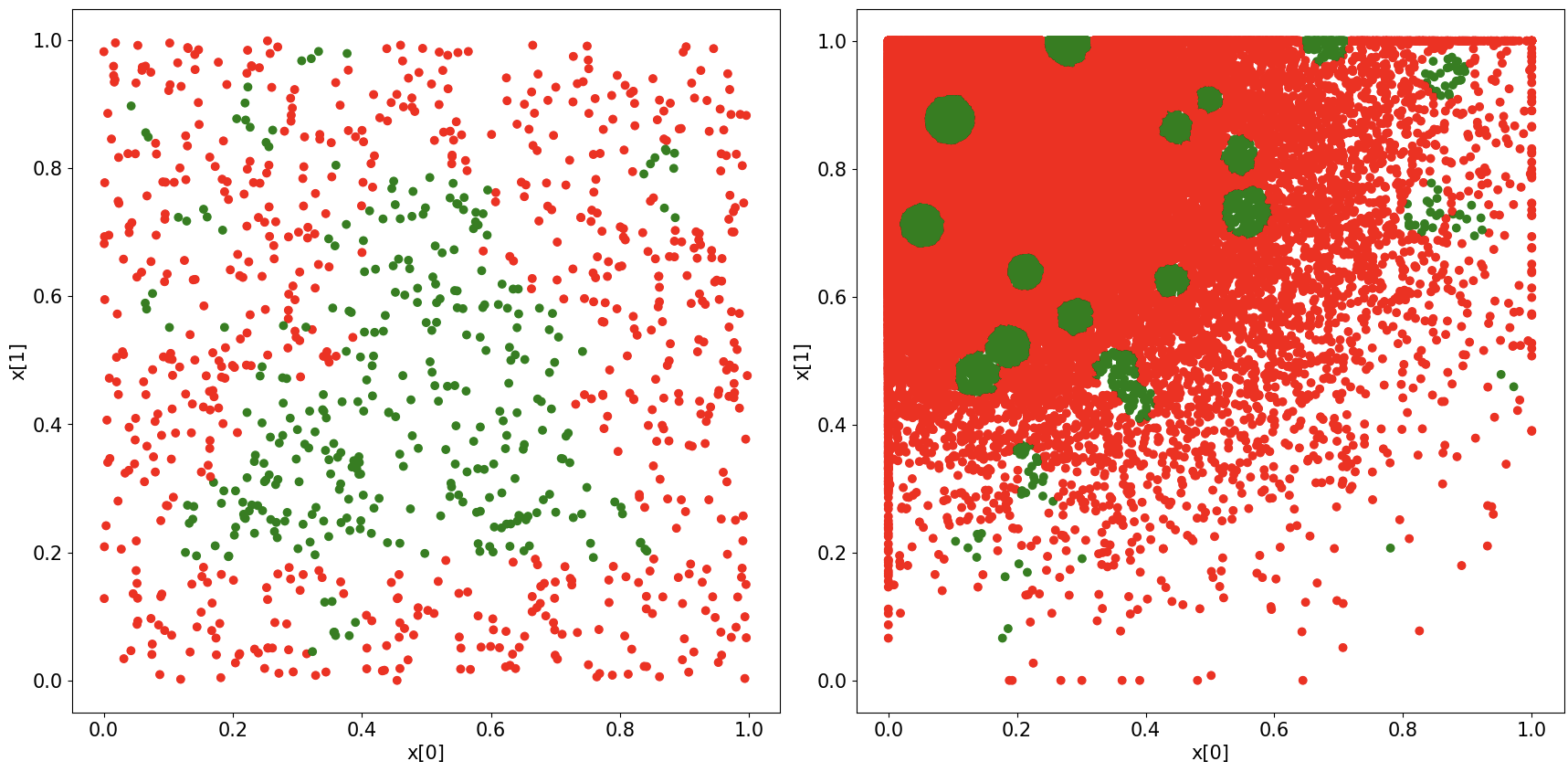}
	\caption{Synthetic datasets DS-1 (lhs) and DS-2 (rhs) representing relatively sparse and dense training data respectively.}
	\label{fig_two_extra_datasets}
\end{figure}

In addition to the running example, we conduct experiments on two synthetic datasets as shown in Figure~\ref{fig_two_extra_datasets}, representing the scenarios with sparse and dense training data respectively. All modelling details and results after applying our RAM on those three datasets are summarised in Table \ref{table_model_details}, based on which we compare the testing error, Average Cell Unastuteness (ACU) and our RAM results ($\myExp[\lambda]$ and $Ub_{97.5\%}$).

\label{sec_evaluation}
\begin{table*}[h]
\centering
\caption{The RAM details and results. For image datasets, the $r$, $\epsilon$ and $\#$ are associated with latent spaces. Time is in seconds per cell.}
\resizebox{\textwidth}{!}{
\begin{tabular}{c|c|c|c|c|c|c|c|c|c}
\hline
            &   train/test error  & $r$-separation & cell radius $\epsilon$ & \# of cells & ACU& $\myExp[\lambda]$ & $\myVar[\lambda]$ & $Ub_{97.5\%}$ & time \\ \hline\hline
The run. exp. &  0.0005/0.0180  &   0.004013          &      0.004       &    $250\times250$    & 0.002982 &   0.004891       &          0.000004      & 0.004899  & 0.04       \\
Synth. DS-1 & 0.0037/0.0800  &  0.004392 &      0.004      &     $250\times250$   &   0.008025   &  0.008290    &      0.000014       &     0.008319 &  0.03 \\
Synth. DS-2 &  0.0004/0.0079&     0.002001       &         0.002               &        $500\times500$       &   0.004739  &   0.005249          &           0.000002        &       0.005252   &  0.04  \\ \hline
MNIST    &  0.0051/0.0235   & 0.1003          &      0.100          &        top-170000       &  0.106615     &     0.036517       &           /        &        /  &   0.43 \\
CIFAR10       &     0.0199/0.0853    &    0.1947   &          0.125              &       top-23000        &       0.238138       &   0.234419  &       /            &        / & 6.74      \\ \hline
\end{tabular}}
\label{table_model_details}
\end{table*}

In the running example, we first observe that the ACU is much lower than the test error, meaning the underlying DL model is a robust one. Since our RAM is mainly based on the robustness evidence, its results are close to ACU but not exactly the same because of the nonuniform OP, cf.\ Figure~\ref{fig_running_example}~(rhs). Moreover, from Figure~\ref{fig_running_example}~(lhs), we know the classification boundary is near the middle of the unit square input space where misclassifications tend to happen (say ``buggy area''), which is also the high density area on the OP. Thus, the contribution to unreliability from the ``buggy area'' is weighted higher by the OP, which explains why our RAM results are worse than ACU.
In contrast, because of the ``flat'' OP in the DS-1 (cf.\ Figure~\ref{fig_two_extra_datasets}~(lhs)), our RAM results are very close to the ACU. With more dense data in DS-2, the $r$-distance is much smaller and leads to more cells. Thanks to the rich data in this case, all three results are more consistent.
We note that, given the nature of the three 2D-point datasets, DL models trained on them are much more robust than image datasets. This is why all ACUs are better than test errors, and our RAM finds a middle point representing reliability according to the OP. Later we apply the RAM on two unrobust DL models trained on image-datasets where the ACUs are worse than test error; it confirms our aforementioned observations.


To gain insights on how to extend our method for high-dimensional/real-world datasets, we also conduct experiments on the popular MNIST and CIFAR10 datasets. Instead of implementing the exact steps in Section~\ref{sec_the_model}, we take a few compromised solutions to tackle the scalability issues raised by ``the curse of dimensionality''. We articulate these steps in the following paragraph, while detailed discussions on their impact on our results are presented in Section~\ref{sec_discussion}.

First, we train Variational Auto-Encoders (VAE) on the MNIST and CIFAR10 datasets and project all inputs into the low dimensional latent spaces of VAE (with 8 and 16 dimensions respectively).
Then we apply the proposed RAM on the compressed dataset, i.e., partitioning the latent space, learning the OP in latent space and evaluating the ``latent-cell unastuteness''.
Astuteness (a special case of robustness), by definition is associated with the input pixel space. By ``latent-cell unastuteness'', we mean the average unastuteness of norm balls (in the input space) around a large number of samples from a ``latent-cell''. The norm ball radius is determined by the $r$-separation distance in the input space.
Taking the computational cost into consideration, we rank the OP of all latent-cells, and choose the top $k$ cells with highest OP for astuteness evaluation. We adopt the existing robustness estimator in \cite{webb_statistical_2019}, where the authors omitted the result of $\myVar[\lambda_i]$; we therefore also omit the variance in our experiments for simplicity. 

\section{Discussions}
\label{sec_discussion}

In this section, we summarise the \textbf{model assumptions} made in our RAM, and discuss if/how they can be validated and what new-assumptions/compromised-solutions are needed to cope with \textbf{high-dimensional/real-world applications}. Finally, we list the \textbf{inherent difficulties} of assessing DL uncovered by our RAM.

\paragraph{Independent $\lambda_i$s and $Op_i$s} As per Assumption \ref{assumption_op_lambda_indep}, we assume all $\lambda_i$s and $Op_i$s are independent when ``assembling'' their estimates via Eq.~\eqref{eq_expected_pfd} and deriving the variance via Eq.~\eqref{eq_varaince_pfd}. Largely this assumption is for the mathematical tractability when propagating the confidence in individual estimates at the cell-level to the whole system \textit{pmi}. Although this independence assumption is hard to justify in practice, it is not unusual in reliability models that use partition, e.g.\ in \cite{pietrantuono_reliability_2020,miller_estimating_1992}. We believe that RAMs are still useful as a first approximation under this assumption, while we envisage that Bayesian estimators leveraging joint priors and conjugacy may relax it.

\paragraph{$R$-separation and its estimation} Assumption \ref{assumption_r_estiamtes_from_data}
derives from Remark \ref{remark_r_sep}. We concur with \cite{yang_closer_2020} and believe that, for any real-world DL classification applications where the inputs are data-points with ``physical meanings'', there should always exist an $r$-stable ground truth.
Such $r$ varies between applications, and the smaller the $r$ is, the harder the inherent difficulty of the classification problem is; i.e., $r$ is a \textit{difficulty indicator} for the given classification problem.

For real-world applications, what really determines the label of an image are its features rather than pixels. Thus, we envisage some latent space (of, e.g., VAE) capturing only the feature-wise information can be explored for high-dimensional data. That is, we
\begin{itemize}
    \item first do $r$-separation based partition in the latent space to learn the OP;
    \item then determine the ground truth labels of cells in the latent space;
    \item map the learned OP and ground truth labels back to the input pixel space;
    \item do astuteness evaluation in the input pixel space and ``assemble'' the results according to the OP.
\end{itemize}


Indeed, it is hard to estimate the $r$ (neither in the input nor the latent space), while the best we can do is to estimate it from the existing dataset. One way of solving the problem is to keep monitoring the $r$ estimates as more labelled data is collected, and redo the cell partition when the estimated $r$ has changed significantly.

\paragraph{Approximation of the OP}
Assumption \ref{assumption_dataset_represents_OP} says that the collected dataset statistically represents the OP, which may not hold for many practical reasons; e.g., the future OP is uncertain at the training stage and thus data is collected in a balanced way to perform well in all categories of inputs. Although we demonstrate our RAM under this assumption for simplicity, it can be easily relaxed. Essentially, we try to fit a density function over the input space from an ``operational dataset'' (representing the OP). Data-points in this set can be \textit{unlabelled} raw data generated from historical data of previous applications, simulations and manually scaled based on expert knowledge. Obtaining such operational dataset is an application-specific engineering problem, and tractable thanks to the fact that it does not require labelled data.

Notably, the OP may also be approximated at runtime based on the data stream of operational data. Efficient KDE for data streams \cite{qahtan_kde_track_2017} can be used. If the OP was subject to sudden changes, change-point detectors like \cite{zhao_interval_2020} should also be paired with the runtime estimator to robustly approximate the OP.

However, we may encounter technical challenges when fitting the PDF from high-dimensional real-world datasets. There are two known major challenges when applying \textit{multivariate} KDE to high-dimensional data: i) the choice of bandwidth $H$ represents the covariance matrix that mostly impacts the estimation accuracy; ii) scalability issues in terms of storing intermediate data structure (e.g.\ data-points in hash-tables) and querying times made when estimating the density at a given input. For the first challenge, the optimal calculation of bandwidth matrix can refer to some rule of thumb \cite{silverman1986density,scott2015multivariate} and the cross-validation \cite{bergstra_random_2012}. While there are dedicated research on improving the efficiency of multivariate KDE, e.g., \cite{backurs2019space} presented a framework for multivariate KDE in provably sub-linear query time with linear space and linear pre-processing time to the dimensions.



\paragraph{Determination of the ground truth of a cell}

Assumptions \ref{assumption_single_gt_cell} and \ref{assumption_empty_cell_label} are essentially on how to determine the ground truth label for a given cell, that relates to the oracle problem of testing DL \cite{guerriero_reliability_2020}. While it is still challenging, we partially solve it by leveraging the $r$-separation property. 

Thanks to $r$, it is easy to determine a cell's ground truth when we see it contains labelled data-points. However, for an empty cell, it is non-trivial. We assume the overall performance of the DL model is fairly good (e.g., better than a classifier doing random classifications), thus miss-classifications within an empty cell are relatively rare events.
Then we can determine the ground truth label of the cell by majority voting of predictions. Indeed, this is a strong assumption when there are some ``failure regions'' in the input space that perform really badly (even worse than random labelling).
In this case, we need to invent a new mechanism to detect such ``really bad failure regions'' and spend more budget on invoking, say, humans to do the labelling. 

\paragraph{Conditional OP of a cell}
We assume the distribution of inputs (i.e., the conditional OP) within each cell is uniform by Assumption \ref{assumption_conditonal_OP_uniform}.
Although we conjecture that this is the common case due to the small size of cells (i.e., those very close/similar inputs within a small region are only subject to noise factors that can be modelled uniformly), the real situation may vary; this requires justification in safety cases. 

For a real-world dataset, the conditional OP represents certain distributions of ``natural variations'' \cite{zhong_understanding_2021}, e.g.\ lighting conditions, obey certain distributions. The conditional OP of cells should faithfully capture the distribution of such natural variations. Recent advance on measuring the natural/realistic AEs \cite{harel_canada_is_2020} highly relates to this assumption and may relax it.

\paragraph{Explosion of the number of cells}
The number of cells to evaluate the astuteness is exponential in the dimensions of data. 
For high-dimensional data, it is impossible to explore all cells in the input space\footnote{Although dimension reduction methods like VAE may ease the problem of learning OP, they cannot reduce the number of cells to be evaluated. Since robustness by definition has to be evaluated in the input space.} as we did for the running example. 
 
A compromised solution is to find the first $k$ cells that \textit{dominate} the OP. That is, we rank the cells by their pooled OP, and only evaluate the top-$k$ cells where the sum of these $k$ cells' OPs is greater than a threshold, e.g.\ 99\%. Then, we can conservatively set the cell \textit{pmi} of the rest 
to a worst-case bound (e.g.~1) or an empirical/average bound based on the first $k$ cells. 
Certainly, the price to pay is to sacrifice estimation accuracy.
The best we can do for now is to increase the budgets for a larger $k$.
Technically, finding the first $k$ cells dominating the OP is in fact to calculate the modes of the KDE function.
The work of \cite{lee2019finding} gives us a hint on how to quickly calculate the modes of Gaussian KDE when the data dimension is high. 

This discussion relates to the cost of our RAM, thus a pertinent question is---what is the real cost of conducting DL testing? Is it the the human labour generating labels or timing constraints? A likely answer is: both. Our RAM has partially solved the former (cf.\ earlier discussions), while the latter is less costly nowadays and can be solved by harnessing the fast growth of computational power and parallel computing.

\paragraph{Efficiency of cell robustness evaluation}

We have demonstrated via the Simple Monte Carlo method to evaluate cell robustness in the running example. It is well-known that Simple Monte Carlo is not a computationally efficient technique to estimate rare-events (such as AEs in our case) in high-dimensional space. Thus, instead of applying Simple Monte Carlo, the more advanced and efficient sampling approach, the Adaptive Multi-level Splitting method \cite{webb_statistical_2019}, has been applied in our case studies on image datasets. We are confident that other statistical sampling methods designed for rare-events may also suffice our need.



In addition to the statistical approach, formal method based verification techniques can also be applied to assess a cell's \textit{pmi}, e.g.\ \cite{huang_safety_2017}. They provide formal guarantees on whether the DL model will miss-classify any input inside a small region. Although such ``robust region'' proved by formal methods is normally smaller than our cells, the $\hat{\lambda}_i$ can be conservatively set to the proportion of robust region covered in $c_i$ in this case.

We would like to note that the cell robustness estimator in our RAM works in a ``hot-swappable'' manner: any new and more efficient robustness estimator can be easily incorporated. Thus, how to improve the efficiency of cell's robustness estimation is out of the scope of our RAM.

\paragraph{Inherent difficulties} Finally, based on our RAM and the discussions above, we summarise the inherent difficulties of assessing DL reliability as the following questions:
\begin{itemize}
    \item How to accurately build the OP in the high-dimensional input space? 
    \item How to build an accurate oracle leveraging the existing human-labels in the training dataset?
    \item What is the local distribution (conditional OP) over a small input region that captures the natural variations of physical conditions? 
    \item How to efficiently evaluate the robustness of a small region given AEs are rare events?
    \item How to sample small regions from a large population (high-dimensional space) to test robustness in an unbiased and efficient way? 
\end{itemize}

We try to provide preliminary/compromised solutions in our RAM, while the questions are still challenging in practice. We doubt the existence of other DL RAMs with weaker assumptions achieving the same level of rigorousness as ours, at this stage.

\section{Conclusion \& Future Work}
\label{sec_conclusion}

In this paper, we present a preliminary RAM for DL classifiers. It is the first DL RAM explicitly considers both the OP information and robustness evidence. It uncovers some inherent difficult questions when assessing DL reliability, while preliminary/compromised solutions are discussed, implemented and demonstrated with case studies. 

An intuitive way of perceiving our RAM, comparing with the usual accuracy testing, is that we enlarge the testing dataset with more test cases around ``seeds'' (original data-points in the test set). We determine the oracle of a new test case according to its seed's label and the $r$-distance. Those enlarged test results form the robustness evidence, and how much they contribute to the overall reliability is proportional to its OP. Consequently, \textit{exposing to more tests (robustness evaluation) and being more representative of how it will be used (the OP)}, our RAM is more trustworthy.

In line with the gist of our RAM, we believe the DL reliability should follow the conceptualised equation of:
$$
DL \,\,\, reliability = generalisability  \times robustness.
$$
In a nutshell, when assessing the DL reliability, we should not only concern how it generalises to a new data-point (according to the future OP), but also the local robustness around it. Align with this insight, indeed, a ``naive/over-simplified'' version of our RAM would be averaging all local astuteness of data-points in the test set, which is less rigorous (e.g., on determining the norm ball size) and requires stronger assumptions (e.g., the test set is equal to the operational set).


Improving the scalability of our RAM and experimenting with more real-world datasets form important future work. We presume a trained DL model for our assessment purpose. A natural question next is how to actually improve the reliability when our RAM results are not good enough. As described in \cite{zhao_detecting_2021}, we plan to investigate DL debug testing (e.g. \cite{huang2021coverage}) and retraining methods \cite{bai_recent_2021}, together with the RAM, to form a closed loop of debugging-improving-assessing.

\section*{Acknowledgments \& Disclaimer}
\includegraphics[height=8pt]{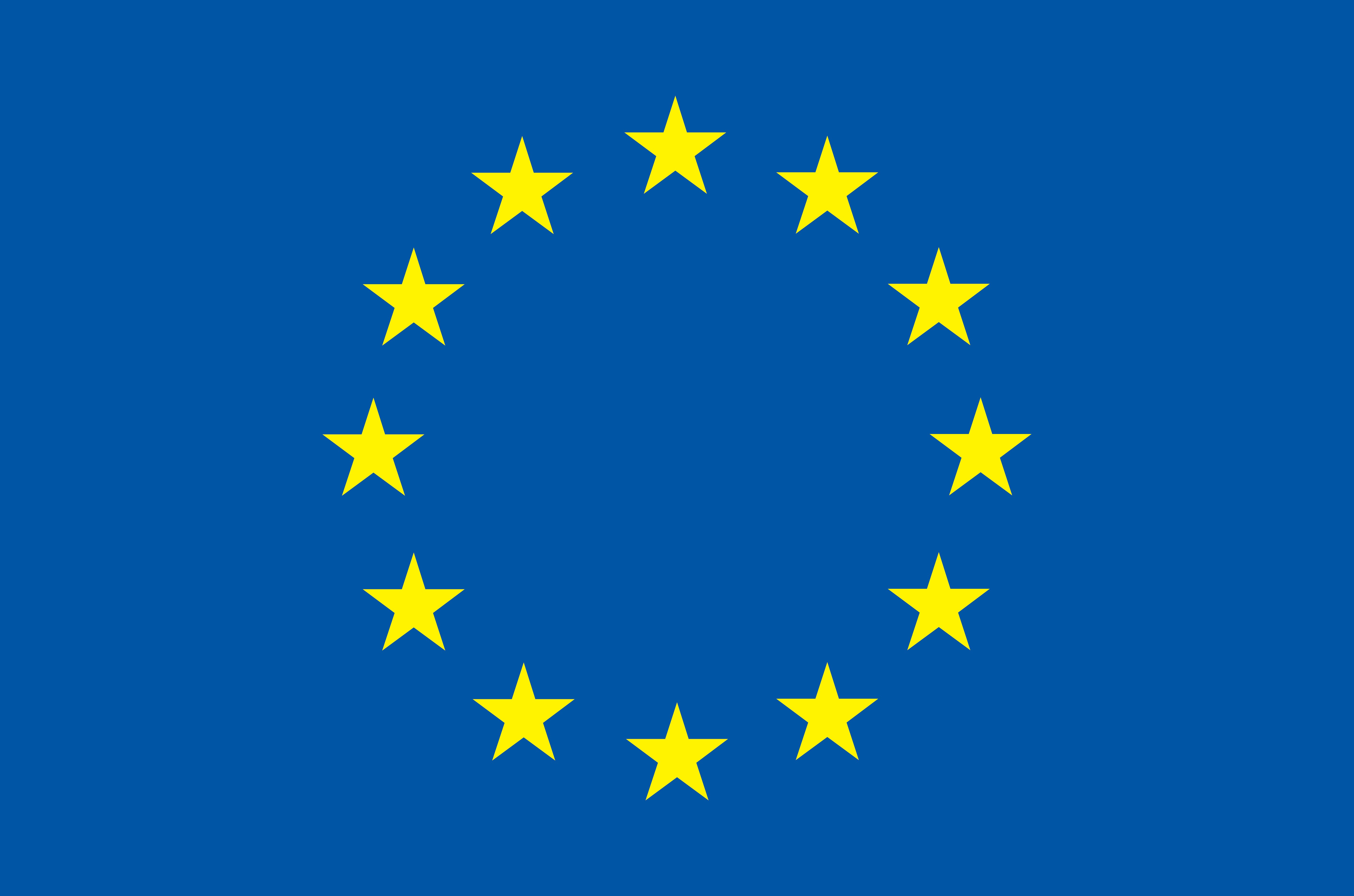} 
This project has received funding from the European Union’s Horizon 2020 research and innovation programme under grant agreement No 956123. 
This work is partially supported by the UK EPSRC (through the Offshore Robotics for Certification of Assets [EP/R026173/1] and End-to-End Conceptual Guarding of Neural Architectures [EP/T026995/1]) and the UK Dstl (through the project of Safety Argument for Learning-enabled Autonomous Underwater Vehicles). Xingyu Zhao and Alec Banks’ contribution to the work is partially supported through Fellowships at the Assuring Autonomy International Programme. We thank Lorenzo Strigini for insightful comments on earlier versions of the paper.

This document is an overview of UK MOD (part) sponsored research and is released for informational purposes only. The contents of this document should not be interpreted as representing the views of the UK MOD, nor should it be assumed that they reflect any current or future UK MOD policy. The information contained in this document cannot supersede any statutory or contractual requirements or liabilities and is offered without prejudice or commitment. Content includes material subject to \textcopyright~Crown copyright (2018), Dstl. This material is licensed under the terms of the Open Government Licence except where otherwise stated. To view this licence, visit \url{http://www.nationalarchives.gov.uk/doc/open-government-licence/version/3} or write to the Information Policy Team, The National Archives, Kew, London TW9 4DU, or email: psi@nationalarchives.gsi.gov.uk.

\appendix
\section{KDE bootstrapping}
Bootstrapping is a statistical approach to estimate any sampling distribution by random sampling method. We sample with replacement from the original data points $(X,Y)$ to obtain a new bootstrap dataset $(X^b, Y^b)$ and train the KDE on the bootstrap dataset. Assume the bootstrap
process is repeated $B$ times, leading to $B$ bootstrap KDEs, denoted as
$\widehat{Op}^1(x), \dots, \widehat{Op}^B(x)$.
Then we can estimate the variance of $\hat{f}(x)$ by the sample variance of the bootstrap KDE \cite{chen2017tutorial}:
$$
\label{eq_bootstrap_var}
\hat{\sigma}_B^2(x) = \frac{1}{B-1} \sum_{b = 1}^B (\widehat{Op}^b(x) - \mu_B)^2
$$
where the $\mu_B$ can be approximated by
$$
\hat{\mu}_B(x) = \frac{1}{B} \sum_{b = 1}^B \widehat{Op}^b(x).
$$

\bibliographystyle{named}
\bibliography{references.bib}

\end{document}